\pdfoutput=1

\documentclass[numbers]{article}

\usepackage[preprint]{neurips_2025}

\usepackage{times}
\usepackage{latexsym}
\usepackage{subcaption}
\usepackage[T1]{fontenc}

\usepackage[utf8]{inputenc}
\usepackage{natbib}
\usepackage{microtype}
\usepackage{booktabs} 
\usepackage{inconsolata}

\usepackage{algorithm}
\usepackage{algpseudocode}
\usepackage{longtable}
\usepackage{listings}
\usepackage[colorlinks,linkcolor=black]{hyperref}
\usepackage{bbding}
\usepackage[utf8]{inputenc} 
\usepackage[T1]{fontenc}    
\usepackage{hyperref}       
\usepackage{url}            
\usepackage{booktabs}       
\usepackage{amsfonts}       
\usepackage{nicefrac}       
\usepackage{microtype}      
\usepackage{xcolor}         
\usepackage{graphicx}
\usepackage{multirow}
\usepackage{array}
\usepackage{colortbl}
\usepackage{enumitem}
\usepackage{wrapfig}
\usepackage{amssymb}
\usepackage{xcolor}         
\usepackage{graphicx}
\usepackage{multirow}
\usepackage{bbding}
\usepackage{array}
\usepackage{colortbl}
\usepackage{enumitem}
\usepackage{tcolorbox}
\tcbuselibrary{breakable}
\usepackage{arydshln}
\usepackage{amsmath}
\title{Mirage-1\includegraphics[width=0.7cm]{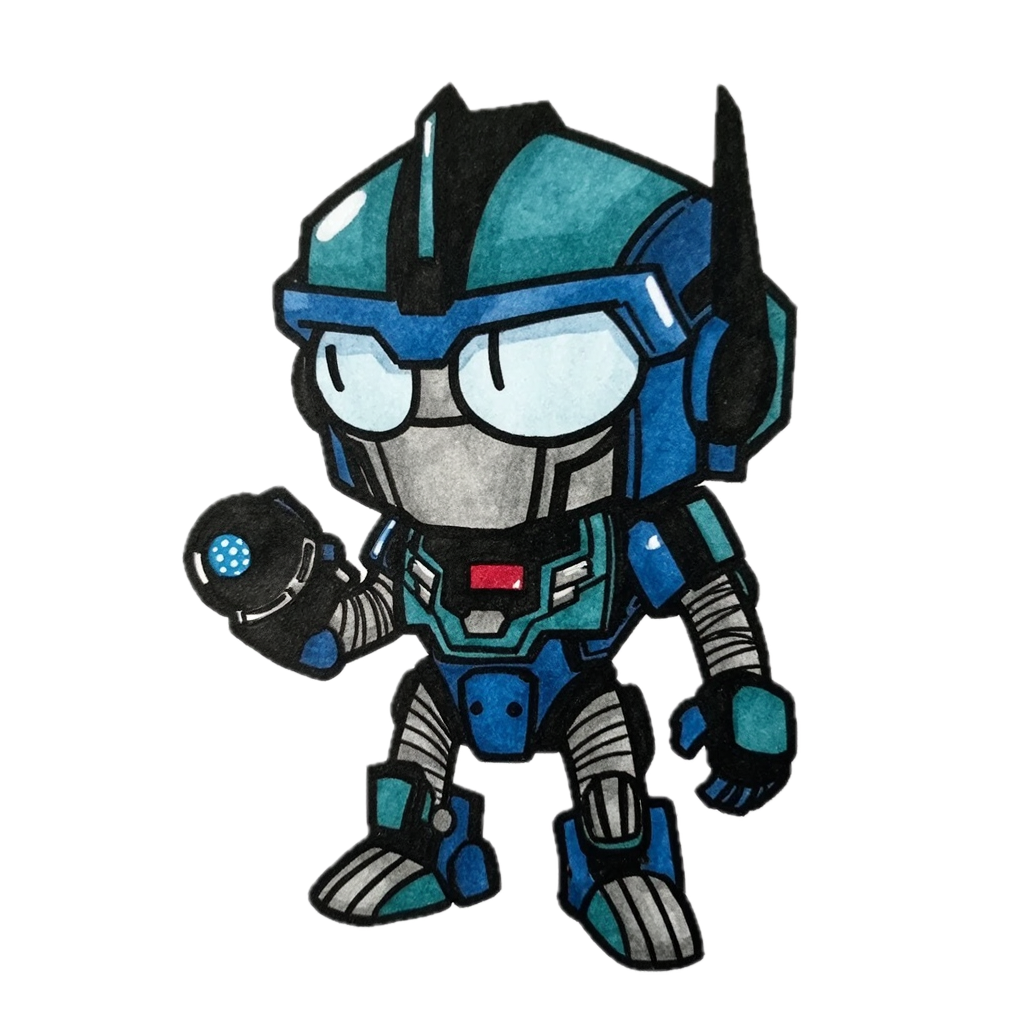}: Augmenting and Updating GUI Agent with Hierarchical Multimodal Skills}

\author{
\parbox{\textwidth}{
    \centering
    \textbf{
Yuquan Xie$^{1}$,
Zaijing Li$^{1\,2}$,
Rui Shao$^{1}$\footnotemark[1], Gongwei Chen$^{1}$ \\  
Kaiwen Zhou$^{3}$, Yinchuan Li$^{3}$, Dongmei Jiang$^{2}$, Liqiang Nie$^{1}$\footnotemark[1]}}
\\ 
    $^1$Harbin Institute of Technology, Shenzhen 
  \hspace{10pt} $^2$Peng Cheng Laboratory \hspace{10pt} $^3$Huawei Noah's Ark Lab\\
       \texttt{\{lzj14011,xieyuquan20016\}@gmail.com, 
       \{shaorui,nieliqiang\}@hit.edu.cn} \\
       \href{https://cybertronagent.github.io/Mirage-1.github.io/}{https://cybertronagent.github.io/Mirage-1.github.io/}
}

\begin{document}
\maketitle
\begin{abstract}
Recent efforts to leverage the Multi-modal Large Language Model (MLLM) as GUI agents have yielded promising outcomes. However, these agents still struggle with long-horizon tasks in online environments, primarily due to insufficient knowledge and the inherent gap between offline and online domains. In this paper, inspired by how humans generalize knowledge in open-ended environments, we propose a \textbf{H}ierarchical \textbf{M}ultimodal \textbf{S}kills (\textbf{HMS}) module to tackle the issue of insufficient knowledge. It progressively abstracts trajectories into execution skills, core skills, and ultimately meta-skills, providing a hierarchical knowledge structure for long-horizon task planning. 
To bridge the domain gap, we propose the \textbf{S}kill-\textbf{A}ugmented \textbf{M}onte \textbf{C}arlo \textbf{T}ree \textbf{S}earch (\textbf{SA-MCTS}) algorithm, which efficiently leverages skills acquired in offline environments to reduce the action search space during online tree exploration. Building on HMS, we propose \textbf{Mirage-1}, a multimodal, cross-platform, plug-and-play GUI agent. To validate the performance of Mirage-1 in real-world long-horizon scenarios, we constructed a new benchmark, AndroidLH. Experimental results show that Mirage-1 outperforms previous agents by 32\%, 19\%, 15\%, and 79\% on AndroidWorld, MobileMiniWob++, Mind2Web-Live, and AndroidLH, respectively.
\end{abstract}

\section{Introduction}

\begin{wrapfigure}[18]{r}{0.49\textwidth}
\centering
\vspace{-8pt}
\includegraphics[width=0.49\textwidth]{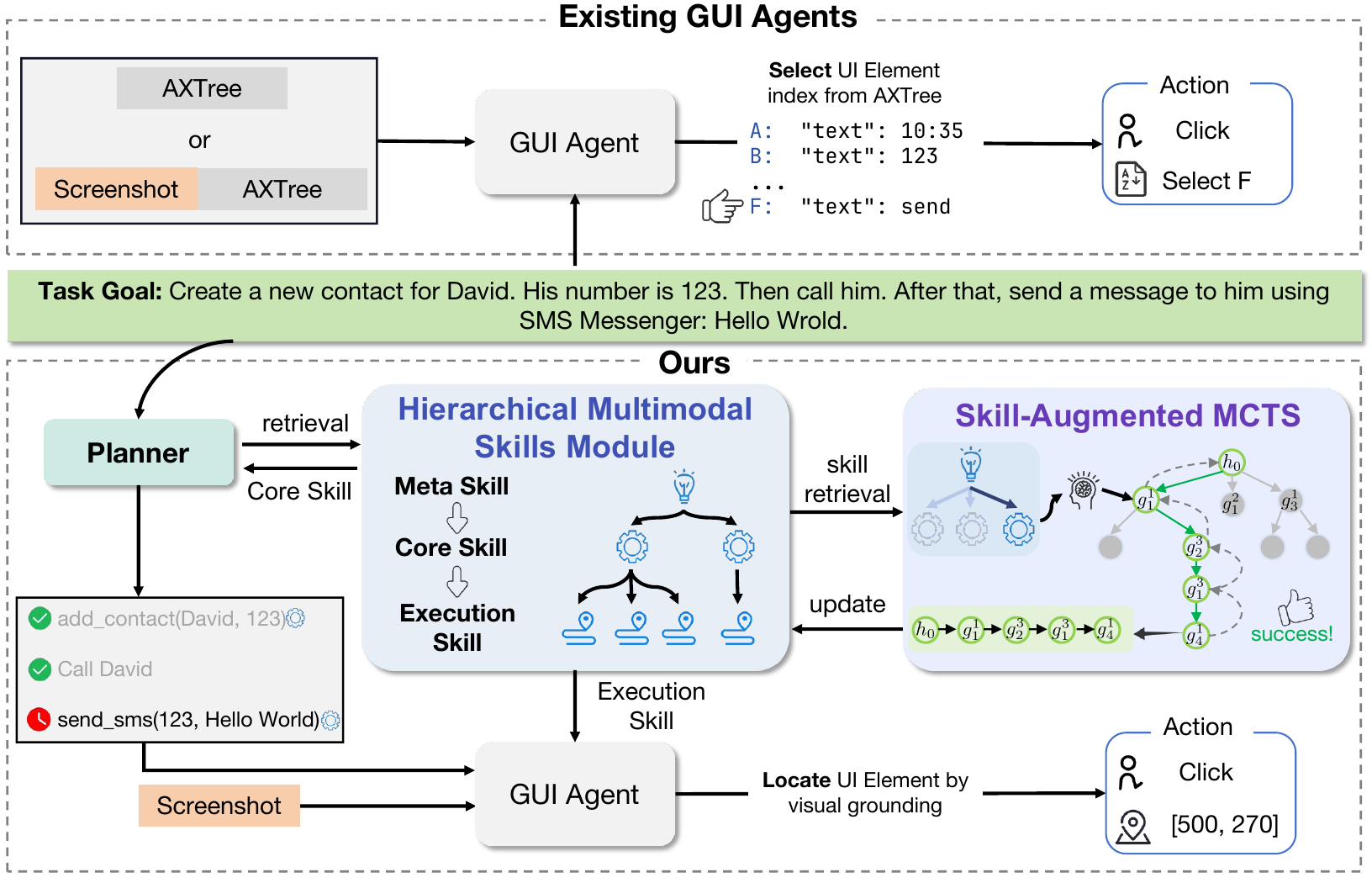}
\vspace{-12pt}
\caption{\small
Compared with existing GUI agents, Mirage-1 improves performance by: (1) \textbf{Hierarchical Multimodal Skills} (HMS) module that enhances the agent's capability in long-horizon task planning, and (2) \textbf{Skill-Augmented MCTS} algorithm that leverages the HMS to search feasible trajectories in online environments.}
\label{fig-intro}
\end{wrapfigure}
The GUI agent that can automate operations on graphical user interfaces, such as mobile devices, represents one of the hallmark applications of artificial general intelligence \cite{hu2024dawn}. Recently, GUI agents driven by Large Language Models (LLMs) have achieved remarkable progress in practical applications \cite{zheng2023synapse,rawles2024androidworld,pan2024webcanvas,zheng2024seeact}. Despite these advancements, these agents \cite{gou2024uground,wu2024osatlas,wang2024mobile-agent-v2} still struggle to execute long-horizon tasks effectively in online environments.

There are two main challenges that arise: \textbf{(1)} Existing GUI agents lack explicit task planning, which limits their ability to perform complex, multi-step operations in dynamic environments. Long-horizon tasks consist of multiple interdependent sub-goals, each demanding that the agent possess sufficient knowledge to grasp the purpose, implementation, and interconnections of these sub-goals. However, as illustrated in Figure \ref{fig-intro}, existing agents make decisions conditioned only on the final goal, lacking fine-grained task planning to guide their actions \cite{wu2024osatlas,gou2024uground,yang2024aria-ui,cheng2024seeclick}.
\textbf{(2)} A domain gap exists between offline and online environments, undermining the agent’s generalization ability. In the online environment, app layouts and button functionalities can change with version updates. Moreover, while apps may share certain similarities, their inherent differences often cause agents to fail when attempting to apply previously acquired knowledge to new applications \cite{zhang2024llm-based-gui-agent-survey,zhang2024guiagent-attacking-popups,grosse2023llm-generalization-with-influence-func,zhang2024oom-generalization-mllm}. As a result, agents trained solely on offline datasets often exhibit subpar performance in online environments \cite{gou2024uground,wu2024osatlas,yang2024aria-ui}.

In this paper, we propose agent \textbf{Mirage-1}, which tackles the aforementioned challenges in two aspects. \textbf{(1) A Hierarchical Multimodal Skills (HMS) module for long-horizon planning}. It first summarizes offline trajectories into \textit{Execution Skills}, then abstracts similar \textit{Execution Skills} into general \textit{Core Skills}, and finally further abstracts them into \textit{Meta Skills}. By clearly outlining the implementation (\textit{Execution Skills}) and functionality (\textit{Core Skills}) of each sub-goal while establishing their interrelationships (\textit{Meta Skills}), HMS provides the comprehensive hierarchical knowledge for long-horizon task planning.
\textbf{(2) A Skill-Augmented Monte Carlo Tree Search (SA-MCTS) algorithm for knowledge exploration in online settings.} To enable the agent to adapt to online environments, we introduce Monte Carlo Tree Search (MCTS) to facilitate exploration and acquisition of online knowledge. Further, we propose Skill-Augmented MCTS (SA-MCTS), which leverages the skills in HMS to perform planning during the search phase. This integration reduce the search space during tree exploration and enhance the reliability of candidate actions to better align with task objectives. Through the SA-MCTS algorithm, the agent quickly acquires skills from the online environment. By integrating and updating skills from both offline and online domains, the domain gap is bridged, leading to improved performance in the online environment. 

We conducted extensive experiments in online environments of mobile and web, and the results substantiate Mirage-1’s superior performance. Compared to current state-of-the-art (SOTA) methods, Mirage-1 achieves average performance gains of 32\%, 19\%, and 15\% on AndroidWorld, MobileMiniWob++, and Mind2Web-Live, respectively. Furthermore, to verify the GUI agent's ability to execute long-horizon tasks in real-world scenarios, we propose an online mobile benchmark, named AndroidLH, which includes 30 long-horizon tasks derived from 12 real-world Android applications. The experimental results show that Mirage-1 achieves a success rate exceeding 79\% on AndroidLH, outperforming previous agents.

In summary, our contributions are as follows:
\begin{itemize}
    \item We propose a cross-platform, plug-and-play GUI agent, Mirage-1. The experimental results demonstrate that Mirage-1 exhibits superior performance on AndroidWorld, MobileMiniWob++, and Mind2Web-Live.  Moreover, on our proposed long-horizon tasks benchmark, AndroidLH, Mirage-1 also achieved the best performance.
    \item To broaden the knowledge base of general MLLMs within the GUI domain, we propose Hierarchical Multimodal Skills (HMS). Summarizing historical trajectories into hierarchical skills efficiently enriches the skill of GUI agents, ultimately leading to a significant improvement in online long-horizon tasks.
    \item To address the domain gap between offline and online environments, we introduce the SA-MCTS algorithm. It leverages the HMS along with the MCTS algorithm to explore unfamiliar tasks, thereby rapidly expanding Mirage-1’s knowledge in online environments.
\end{itemize}

\section{Related Work}
\subsection{GUI Agents}
\begin{figure*}[htbp]
    \centering
    \includegraphics[width=1.0\textwidth]{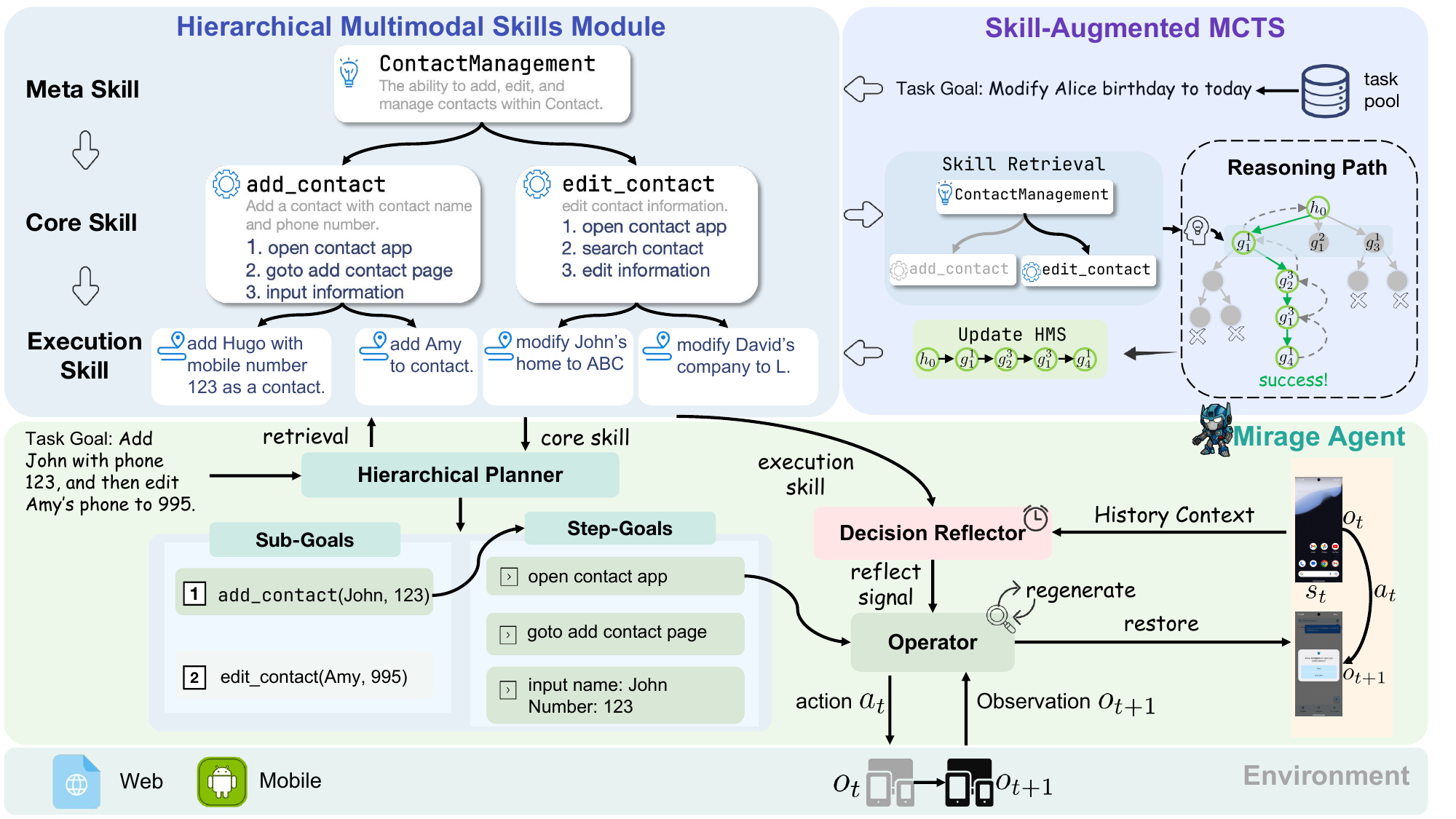}
    \caption{The Mirage-1 framework comprises a Hierarchical Planner, an Operator, a Decision Reflector, and a Hierarchical Multimodal Skills Module (HMS). To bridge the offline-online domain gap, Skill-Augmented Monte Carlo Tree Search (SA-MCTS) is employed for unseen task exploration, with successful trajectories expanding HMS capabilities. The Hierarchical Planner retrieves Core Skills from HMS and decomposes task goals into sub-goals for Operator execution. The Decision Reflector leverages Execution Skills to assess task execution feasibility.}
    \label{fig-framework}
\end{figure*}

Graphical User Interface (GUI) agents are designed to autonomously interact with digital systems through the emulation of human-like interactions, including clicking and typing. Previous approaches to GUI automation have predominantly relied on HTML or Accessibility Trees for element grounding and action execution \cite{zheng2023synapse,rawles2024androidworld,lai2024autowebglm,deng2024mind2web,gur2023webagent,zheng2024seeact,xie2025gui}. However, there are differences between this approach and the way humans use digital devices. Recently, several studies \cite{wu2024osatlas,gou2024uground,yang2024aria-ui,lu2024omniparser,xu2024aguvis,cheng2024seeclick,ge2024iris,lin2024showui,wang2024mobile-agent-v2} have proposed the development of a GUI agent based on pure vision. These approaches utilize a multimodal planner to plan the current task and generate actions, alongside a grounding model to locate elements. Despite their effectiveness on offline datasets, these approaches exhibit limitations in completing long-horizon tasks within complex and dynamic online environments.
Mirage-1 addresses long-horizon task planning by integrating HMS, which provides hierarchical knowledge. Mirage-1 utilizes the SA-MCTS algorithm to explore unknown tasks in an online environment, thereby reducing the domain gap between offline and online settings.

\subsection{Memory in Agents}

In agent-environment interactions, memory mechanisms are crucial for planning, reasoning, and reflection. Recent studies have focused on textual memory representations for their interpretability and efficiency \cite{zhang2024memory_agent_survey}. Notable approaches include Agent Workflow Memory \cite{wang2024workflow-memory}, leveraging recurring patterns for planning, Synapse \cite{zheng2023synapse} utilizing abstracted state-action trajectories, and Agent S \cite{agashe2024agent-s} incorporating task experiences in hierarchical planning. Despite their effectiveness, text-based approaches exhibit limitations in processing multimodal information. Although \cite{li2024optimus1} introduced multimodal memory in Optimus-1, its graph-based representation faces scalability challenges in GUI environments. 
These limitations are addressed by the hierarchical multimodal skills module in Mirage-1, which progressively abstracts historical trajectories into structured, hierarchical skills. The hierarchical multimodal skills module provides comprehensive hierarchical knowledge for long-horizon task planning.

\section{Mirage-1}
In this section, we first define the task formulation, then introduce the framework of Mirage-1, followed by an in-depth overview of \textbf{H}ierarchical \textbf{M}ultimodal \textbf{S}kills (\textbf{HMS}), and finally introduce the \textbf{S}kill-\textbf{A}ugmented \textbf{M}onte \textbf{C}arlo \textbf{T}ree \textbf{S}earch (\textbf{SA-MCTS}) algorithm.

\subsection{Task Formalization}

Formally, we denote the Environment as $\mathcal{E}$, the Observation Space as $\mathcal{O}$, the Action Space as $\mathcal{A}$, and the Multimodal Agent as $\mathcal{M}$.
Given a task goal $G$, at time step $t$, the agent receives the observation $o_t$ and context $C_t$ as input. $C_t$ consists of historical actions $a_i$ and summary $s_i$, defined as: $C_t=\left \{ g,a_1,s_1,\dots,a_{t-1},s_{t-1} \right \} $. 
The agent predicts the action $a_t$ and its corresponding description $d_t$ at time $t$, where $\left \{ a_t, d_t \right \} = \mathcal{M}(o_t, C_t)$. 
The action $a_{t}$ interacts with the environment $\mathcal{E}$, resulting in the observation $o_{t+1}$ at the next time step. 
Through continuous interactions, we obtain a trajectory \( J =\) \(\{(o_1, a_1), (o_2, a_2), (o_3, a_3), \ldots, (o_T, a_T)\}\), where \( T \) represents the length of the trajectory. 

\subsection{Framework of Mirage-1}

As illustrated in Figure \ref{fig-framework}, Mirage-1 consists of four modules: Hierarchical Planner, Operator, Decision Reflector, and Hierarchical Multimodal Skills Module.

\noindent\textbf{Hierarchical Planner.}
Given a task goal $G$, the Hierarchical Planner $\mathcal{A}_P$ first retrieves the candidate Meta Skills from HMS by cosine similarity calculation:
\begin{equation}
m_1,\dots,m_k \gets  CoSim(G,m_i)
\end{equation}
where \(CoSim(\cdot)\) denotes the cosine similarity, $m_i$ represents the Meta Skills in HMS, and $k$ indicates the number of candidates. Subsequently, $\mathcal{A}_P$ evaluates these candidates to determine the most appropriate Meta Skill $m_c$:
\begin{equation}
m_c=\mathcal{A}_P(G, m_1,\dots,m_k)
\end{equation}

Finally, $\mathcal{A}_P$ leverages the Core Skills associated with $m_c$ as conditions to generate a sequence of sub-goals:
\begin{equation}
g_1,g_2,\dots,g_n = \mathcal{A}_P(G, c_1,\dots,c_r)
\end{equation}
where $r$ and $n$ indicate the number of Core Skills and the number of sub-goals, respectively.
We implement $\mathcal{A}_P$ using \texttt{GPT-4o} \cite{hurst2024gpt-4o} as the reasoning model and \texttt{text-embedding-3-small} for cosine similarity computations.

\noindent\textbf{Operator.} Given the current sub-goal $g_i$ generated by $\mathcal{A}_P$, along with observation $o_t$ and historical summary $S_{1 \dots t}$ at step $t$, the Operator $\mathcal{A}_O$ generates an action $a_t$ and its description $d_t$:
\begin{equation}
    a_t, d_t=\mathcal{A}_O(g_i,o_t, S_{1 \dots t})
\end{equation}

For interactive actions such as click, double-click, long press, text input, and scroll, $\mathcal{A}_O$ invokes a visual grounding model to locate the target element coordinates:
\begin{equation}
    \mathrm{coord}_t = \mathcal{A}_O(d_t,o_t)
\end{equation}

For predefined actions such as scrolling up or returning to a previous screen, $\mathcal{A}_O$ directly invokes the corresponding system API to execute the action. For both mobile and web environments, we implement our Operator by combining agents with grounding models. Specifically, we use M3A Agent \cite{rawles2024androidworld} for mobile environments and WebCanvas Agent \cite{pan2024webcanvas} for web environments. To demonstrate the generalization of the proposed method, we employ various grounding models, including OS-Atlas \cite{wu2024osatlas}, UGround \cite{gou2024uground}, Aria-UI \cite{yang2024aria-ui}, and UI-TARS \cite{qin2025ui-tars}.

\noindent\textbf{Decision Reflector.}
The GUI environment presents unique challenges where each action can significantly impact subsequent states. The Decision Reflector $\mathcal{A}_R$ addresses this challenge by assessing the effectiveness of the action and preventing potential state conflicts.

Operating at a fixed frequency, $\mathcal{A}_R$ analyzes each predicted action $a_t$ within its context. It takes the task goal $G$, the current sub-goal $g_i$, observation $o_t$, and historical summary $S_{1 \dots t}$ as input while leveraging the Execution Skills $\varepsilon$ in $\mathrm{HMS}$. $\mathcal{A}_R$ then evaluates potential state changes $p$ and assigns a quality score $q$ (range $0$-$10$):
\begin{equation}
p,q=\mathcal{A}_R(G,g_i,o_t,a_t,S_{1 \dots t}, \varepsilon)
\end{equation}

When $q$ falls below 5, the Operator $\mathcal{A}_O$ receives a regeneration signal to produce an alternative action. Experimental results in Section \ref{sec:ablation} validate that this reflection mechanism substantially improves task success rates in online settings.

\subsection{Hierarchical Multimodal Skills}
\label{sec:hms}
Humans exhibit remarkable generalization capabilities when learning skills across different graphical user interface (GUI) environments, due to they are able to abstract specific tasks into generalizable skills \cite{human-learn-structure}. When faced with a new task, humans draw on previously learned skills as guidance. For example, once a person learns to perform the task \texttt{Search for the weather in London on April 1st}, they can abstract it into the skill, ``search for the weather of a given location and date''. The process typically involves fundamental steps such as launching a browser, entering a query, and executing a search, these steps remain consistent across different systems. Consequently, the person can effortlessly apply this skill to similar tasks in new contexts, such as \texttt{Search for the weather in New York on March 2nd}.


Inspired by this, we propose the \textbf{H}ierarchical \textbf{M}ultimodal \textbf{S}kills (\textbf{HMS}) module, which abstracts specific tasks into generalizable skills and provides structured guidance for task planning. This enables the agent to leverage prior experiences as reusable planning references when encountering new tasks. As shown in Figure \ref{fig-framework}, we divide the Hierarchical Multimodal Skills into three levels: \textit{Meta Skill}, \textit{Core Skill}, and \textit{Execution Skill}.

\textbf{\textit{Execution Skill} represents the summarization and formalization of historical trajectories, which enhances the efficiency and convenience of knowledge development and retrieval during reflection.} To identify the intention (referred to as the `step-goal') behind each action in trajectory $J$, we provide \texttt{GPT-4o} with the temporal sequence of observations ($o_{t-1}$ and $o_t$) encompassing the executed action $a_t$, in conjunction with the current sub-goal $g$. \texttt{GPT-4o} is then instructed to analyze the changes in these observations to infer the corresponding step-goal $s_i$ for the current action $a_t$. Execution Skill is represented as $\varepsilon = (g, o_1, s_1, a_1, \dots, o_{t-1}, s_{t-1}, a_{t-1}, o_t)$. For example, \texttt{Search for the weather in London on April 1st} is an \textit{Execution Skill}, as it specifies a particular date and city. 
            

\begin{algorithm}[t]\small
\caption{Skill-Augmented MCTS}
\label{alg:mcts}
\renewcommand{\algorithmicrequire}{\textbf{Input:}}
\renewcommand{\algorithmicensure}{\textbf{Output:}}
\begin{algorithmic}
\Require $\mathcal{M}$: Multimodal Agent, $\mathcal{D}_T$: task pool, $N$: number of iterations, $T$: MCTS tree depth, $S$: number of sub-goals to sample, $\mathcal{B}$: replay buffer, $\mathcal{K}_0$: hierarchical multimodal skill module initialized by offline datasets
\Ensure $\mathcal{K}_N$, the latest hierarchical multimodal skill module
\For{$i=1$ to $N$} \Comment{\textcolor{red}{Iterative skill module refinement loop}}
    \State $\mathcal{K}_i \gets \mathcal{K}_{i-1}$ \Comment{Inherit skills from previous iteration}
    \For{each task in $\mathcal{D_T}$} \Comment{Process each task to gather experience}
        \State Initialize MCTS tree $\mathcal{TREE}$ with root $h_0$.
        \For{$t=1$ to $T$}
            \State $l_t \gets \text{SelectLeafNode}(\mathcal{TREE}, h_0)$ \Comment{Select leaf node $l_k$}
            \State \textbf{Sub-Goal Expansion:}
            \State \quad $\{g_t^1, \dots, g_t^S\} \gets \text{SampleSubGoals}(\mathcal{K}_i, l_t, S)$ \Comment{\textcolor{red}{Generate $S$ sub-goals via $\mathcal{K}_i$}}
            \State \quad State $\{r_t^1, \dots, r_t^S\} \gets \text{EstimateValues}(\mathcal{M}, l_t, \{g_t^j\}_{j=1}^S)$ \Comment{Evaluate sub-goals}
            \State \quad Select sub-goal $g_t^* = \arg\max_{j} r_t^j$ \Comment{\textcolor{red}{Choose best sub-goal}}
            \State \textbf{Sub-Goal Rollout:}
            \State \quad $(\tau_t,  R_t^{rollout}) \gets \text{ExecuteSubGoal}(\mathcal{M}, \mathcal{K}_i, l_k, g_k^*)$ \Comment{\textcolor{red}{Execute $g_k^*$ using skills}}
            \State \textbf{Backpropagation:}
            \State \quad Update value estimates along the path from $l_t$ to root \Comment{Update $N(\cdot), Q(\cdot)$ from $l_k$ to root}
            \State \quad Store $(g_t^*, \tau_t, R_t^{rollout}$) in node
        \EndFor
        \State Human evaluates whether success or not and adds successful trajectories to the replay buffer $\mathcal{B}$
    \EndFor \Comment{End of task processing loop}
    \State $\mathcal{K}_i \gets \text{RefineSkillModule}(\mathcal{K}_{i-1}, \mathcal{B})$ \Comment{\textcolor{red}{Refine $\mathcal{K}_i$}}
\EndFor \Comment{End of skill refinement iterations}
\end{algorithmic}
\end{algorithm}

\textbf{\textit{Core Skill} is a general function derived from multiple \textit{Execution Skills}.} It is represented as a function, and the agent can obtain step goals for completing a specific task by invoking this function. For example, ``search for the weather of a given location and date'' can be represented as a \textit{Core Skill}: $\mathrm{search\_weather(location, date)}$. We employ \texttt{GPT-4o} to consolidate functionally similar \textit{Execution Skills} into a generalized \textit{Core Skill}. Thus, a \textit{Core Skill} represents a universal pattern for a class of tasks, while the associated \textit{Execution Skills} summarize the specific task trajectories related to it.

\textbf{\textit{Meta Skill} is a high-level aggregation of \textit{Core Skills}.} For example, the \textit{Meta Skill} $\mathrm{search\_web()}$ encompasses \textit{Core Skills} such as $\mathrm{search\_news(query,date)}$, $\mathrm{search\_weather(location, date)}$, and $\mathrm{search\_wiki(query)}$. We employed \texttt{GPT-4o} to consolidate \textit{Core Skills} into \textit{Meta Skills}. Notably, the same \textit{Core Skill} may be applicable to different \textit{Meta Skills}, thereby implicitly establishing the interrelationships between \textit{Meta Skills}.

The HMS module starts empty. Upon encountering a new \textit{Execution Skill}, it is first analyzed at the \textit{Meta Skill} level. If no suitable \textit{Meta Skill} category exists, \texttt{GPT-4o} is utilized to generate a new \textit{Meta Skill}, accompanied by comprehensive documentation. Subsequently, the existing skills associated with this \textit{Meta Skill} are reviewed. \texttt{GPT-4o} evaluates whether these skills sufficiently abstract the task's characteristics. If the existing skills are deemed inadequate, a new \textit{Core Skill} is created and categorized under the appropriate \textit{Meta Skill}. To maintain memory retrieval efficiency, the \textit{Core Skill} layer is regularly refined. Similar \textit{Core Skills} are merged to enhance both retrieval accuracy and the overall representation of knowledge.

\subsection{Skill-Augmented MCTS}
Due to the initial construction of the HMS based on offline trajectories, when deployed in online environments, the agent encounters a significant domain gap. This occurs because offline trajectories often contain outdated information that diverges from the dynamic nature of online content. To address this challenge, we propose a \textbf{S}kill-\textbf{A}ugmented \textbf{M}onte \textbf{C}arlo \textbf{T}ree \textbf{S}earch (\textbf{SA-MCTS}) approach for online exploration. Our HMS module improves the MCTS \cite{hao2023reasnwithlanguagemodel,putta2024agent-q} process by providing generalized planning strategies, while online exploratory techniques simultaneously enrich the skill base. This bidirectional interaction improves both the agent's real-time adaptability and the universal applicability of knowledge.


SA-MCTS is illustrated in Algorithm \ref{alg:mcts}. During exploration, the agent generates sub-goals through a hierarchical process: it first identifies relevant \textit{Meta Skills} for the current task and then selects appropriate \textit{Core Skills} in \textit{Meta Skills} to generate specific sub-goals. This knowledge-guided approach improves both exploration efficiency and the agent's ability to generalize across complex tasks. Successful trajectories are systematically processed: they are first formatted as \textit{Execution Skills}, then abstracted into \textit{Core Skills}, and finally categorized under appropriate \textit{Meta Skills}. 

\definecolor{DarkGreen}{rgb}{0.0, 0.5, 0.0}
\begin{table*}[t]\small
\caption{\label{mobile_result}
Performance comparison on AndroidWorld, MobileMiniWob++, and AndroidLH. $^*$ means the results are reproduced under the same prompt setting.
}
\centering
\resizebox{\textwidth}{!}{
\begin{tabular}{clcccccccc}
\toprule[1.5pt]
\textbf{Method} & \textbf{Operator} & \multicolumn{2}{c}{\textbf{AndroidWorld}} & \multicolumn{2}{c}{\textbf{MobileMiniWob++}} & \multicolumn{4}{c}{\textbf{AndroidLH}}  \\ 
\midrule
 &  & SR & $\Delta$ & SR & $\Delta$ & CR & $\Delta$ & SR & $\Delta$ \\ \midrule
 & \multicolumn{9}{c}{\textbf{AxTree}} \\ \midrule
\multirow{3}{*}{\citeauthor{rawles2024androidworld}} & GPT-4-Turbo+Choice & 30.6 & - & 59.7 & - & 33.3 & - & 13.3 & - \\
 & Gemini 1.5Pro+Choice & 19.4 & - & 57.4 & - & 34.9 & - & 16.7 & - \\
 & GPT-4o+Choice & 41.9 & - & - & - & 38.1 & - & 20.0 & - \\ \midrule
 & \multicolumn{9}{c}{\textbf{Image+AXTree}} \\ \midrule
\multirow{3}{*}{\citeauthor{rawles2024androidworld}} & GPT-4-Turbo + SoM & 25.4 & - & 67.7 & - & 34.9 &  -& 16.7 &-  \\
 & Gemini 1.5 Pro + SoM & 22.8 & - & 40.3 & - & 36.5 & - & 20.0 &-  \\
 & GPT-4o + SoM & 36.6 &-  & - &  -& 31.7 & - & 23.3 & - \\
\citeauthor{zheng2024seeact} & GPT-4-Turbo + SoM & 15.5 & - & 66.1 & - & 30.2 & - & 13.3 &-  \\ \midrule
 & \multicolumn{9}{c}{\textbf{Image}} \\ \midrule
\citeauthor{wu2024osatlas} & GPT-4o + OS-Atlas & 31.9$^*$ & - & 51.1$^*$ & - & 27.0 & - & 16.7 &-  \\
\rowcolor[HTML]{E7EEFE} Mirage-1-O &GPT-4o + OS-Atlas  & 42.2$^*$ & \textcolor{DarkGreen}{$\uparrow \textbf{32.3}\%$} & 60.9$^*$ & \textcolor{DarkGreen}{$\uparrow \textbf{19.1}\%$} & 50.8 & \textcolor{DarkGreen}{$\uparrow \textbf{88.1}\%$} & 30.0 & \textcolor{DarkGreen}{$\uparrow \textbf{79.6}\%$} \\
 \citeauthor{gou2024uground} & GPT-4o + UGround & 32.8$^*$ & - &  48.4$^*$& - &  39.7& - & 26.7 & - \\
\rowcolor[HTML]{E7EEFE} Mirage-1-U &GPT-4o + UGround  & 40.5$^*$ &  \textcolor{DarkGreen}{$\uparrow \textbf{23.4}\%$}& 56.5$^*$ & \textcolor{DarkGreen}{$\uparrow \textbf{16.7}\%$} & 55.6 & \textcolor{DarkGreen}{$\uparrow \textbf{40.1}\%$} &36.7  & \textcolor{DarkGreen}{$\uparrow \textbf{37.4}\%$} \\ 
\citeauthor{yang2024aria-ui}  & GPT-4o + Aria-UI & 37.1$^*$ & - & 52.2$^*$ & - & 31.7 & - & 23.3 & - \\
\rowcolor[HTML]{E7EEFE} Mirage-1-A &GPT-4o + Aria-UI  & 41.4$^*$ &\textcolor{DarkGreen}{$\uparrow \textbf{11.6}\%$}  &62.0$^*$  & \textcolor{DarkGreen}{$\uparrow \textbf{18.8}\%$} & 49.2 & \textcolor{DarkGreen}{$\uparrow \textbf{55.2}\%$} & 36.7 & \textcolor{DarkGreen}{$\uparrow \textbf{57.5}\%$} \\ 
\citeauthor{qin2025ui-tars}  & GPT-4o + UI-TARS & 36.2$^*$ & - & 53.3$^*$ & - & 42.9 & - & 26.7 &-  \\
\rowcolor[HTML]{E7EEFE} Mirage-1-T &GPT-4o + UI-TARS  & 43.1$^*$ &  \textcolor{DarkGreen}{$\uparrow \textbf{19.1}\%$}& 60.9$^*$ & \textcolor{DarkGreen}{$\uparrow \textbf{14.3}\%$} & 55.2 & \textcolor{DarkGreen}{$\uparrow \textbf{28.6}\%$} & 40.0 & \textcolor{DarkGreen}{$\uparrow \textbf{49.8}\%$}  \\ 
 \bottomrule[1.5pt]
\end{tabular}
}
\vspace{-5pt}
\end{table*}

In summary, the SA-MCTS algorithm enhances MCTS exploration by incorporating offline-initialized HMS in two key aspects. First, instead of directly generating actions, the agent utilizes existing \textit{Meta Skills} and \textit{Core Skills} to decompose the task goal into sub-goals. Second, these skill-guided sub-goals direct the agent towards more promising exploration paths, significantly reducing ineffective searches. See Appendix \ref{appendix-sa-mcts} for more details.

\section{Experiments}

\subsection{Experimental Settings}
We evaluate the proposed agent, Mirage-1, in both \textbf{Mobile} and \textbf{Web} online environments to demonstrate its cross-platform generalization. For Mobile, we evaluate Mirage-1's performance on AndroidWorld \cite{rawles2024androidworld}, an online benchmark within an Android emulator environment. We report \textit{task success rate} as an evaluation metric. For Web, we employ Mind2Web-Live \cite{pan2024webcanvas} as the evaluation benchmark. We use \textit{task success rate} and \textit{micro-completion rate} (measures the proportion of key nodes completed in all tasks) as evaluation metrics. For more details, please refer to Appendix \ref{appendix-exp-setting}.

\subsection{Baselines}
We conducted a comprehensive comparison with various baselines in mobile and web environments.

\noindent\textbf{Mobile.} We compare the baselines of three input modes: AXTree-only, image-AXTree, and image-only. For AXTree-only, we employ \texttt{GPT-4-Turbo}, \texttt{Gemini-1.5-Pro} \cite{team2024gemini-1.5} and  \texttt{GPT-4o} as Planner with choice-based selection \cite{rawles2024androidworld}.
For Image-AXTree, we compare \texttt{GPT-4-Turbo}, \texttt{Gemini-1.5-Pro}, \texttt{GPT-4o}, and SeeAct \cite{zheng2024seeact}, all implementing SoM grounding methods \cite{yang2023setofmark}.
For image-only, we employ \texttt{GPT-4o} as the Planner and utilize four grounding models trained in offline environments: OS-Atlas \cite{wu2024osatlas}, UGround \cite{gou2024uground}, Aria-UI \cite{yang2024aria-ui}, and UI-TARS \cite{qin2025ui-tars}. 

\noindent{\textbf{Web.}} For text-only input, we employ the WebCanvas agent \cite{pan2024webcanvas} with various Planner, including \texttt{GPT-4-Turbo}, \texttt{GPT-4o}, \texttt{Gemini-1.5-Pro}, and \texttt{GPT-3.5-Turbo} with choice-based selection. For image-only input, we remove the DOM tree components in the observation. We compare the performance of \texttt{GPT-4o} with OS-Atlas \cite{wu2024osatlas}, UGround \cite{gou2024uground} and Aria-UI \cite{yang2024aria-ui}.
See Appendix \ref{appendix-exp-baseline} for more details.

\subsection{Implementation Details}

\textbf{Mobile.} We build Mirage-1 based on vision-only M3A Agent \cite{rawles2024androidworld}. We used the trajectories in AITW \cite{rawles2024aitw} to build the initial HMS. We prompt \texttt{GPT-4o} to generate 200 unseen tasks and use SA-MCTS to expand HMS. See Appendix \ref{appendix-exp-impl} for more details.

\noindent\textbf{Web.} We build Mirage-1 based on the vision-only WebCanvas agent \cite{pan2024webcanvas}. We use the trajectories of Multimodal-Mind2Web \cite{zheng2024seeact} to build the initial HMS, which is the multimodal version of Mind2Web \cite{deng2024mind2web}. We prompt \texttt{GPT-4o} to generate 200 unseen tasks and use SA-MCTS to update HMS.

\subsection{Experiments Result}
\begin{wraptable}[16]{r}{0.49\textwidth}
\centering
\small
{
\vspace{-12pt}
\caption{\small
Performance comparison on Mind2Web-Live.
We report the completion rate (CR) and the task success rate (SR). $^*$ means the results are reproduced under the same prompt setting.}
\vspace{-5pt}
\label{mind2web_live_result}
{

\resizebox{0.5\textwidth}{!}{%
\begin{tabular}{l|lll}
\toprule[1.5pt]
\textbf{Method} & \textbf{Operator} & \textbf{CR} & \textbf{SR} \\ 
\midrule
 & \multicolumn{3}{c}{\textbf{HTML Tree}} \\ \midrule
\multirow{4}{*}{\citeauthor{pan2024webcanvas}}  & GPT-4-Turbo + Choice & 44.3 & 21.1 \\
& GPT-4o + Choice  & 47.6 & 22.1 \\
& Gemini-1.5-Pro + Choice  & 44.6 & 22.3 \\
& GPT-3.5-Turbo + Choice  & 40.2 & 16.5 \\ 
\midrule
& \multicolumn{3}{c}{\textbf{Image}} \\ 
\midrule

\citeauthor{wu2024osatlas} &GPT-4o + OS-Atlas & 50.4$^*$ & 18.4$^*$ \\
\rowcolor[HTML]{E7EEFE}
Mirage-1-O & GPT-4o + OS-Atlas & \textbf{51.2} & \textbf{21.3}$_{\color{DarkGreen}{\uparrow\textbf{\scriptsize15.8\%}}}$ \\ 
\citeauthor{gou2024uground} &GPT-4o + UGround & 50.8$^*$ & 19.2$^*$ \\
\rowcolor[HTML]{E7EEFE}
Mirage-1-U & GPT-4o + UGround & \textbf{51.4} & \textbf{22.1}$_{\color{DarkGreen}{\uparrow\textbf{\scriptsize15.1\%}}}$ \\ 
\citeauthor{yang2024aria-ui} & GPT-4o + Aria-UI & 51.9$^*$ & 22.1$^*$ \\
\rowcolor[HTML]{E7EEFE}
Mirage-1-A & GPT-4o + Aria-UI & \textbf{53.3} & \textbf{24.0}$_{\color{DarkGreen}{\uparrow\textbf{\scriptsize8.6\%}}}$ \\ 
\bottomrule[1.5pt]
\end{tabular}

}
}
}
\end{wraptable}

\textbf{Mobile.} As illustrated in Table \ref{mobile_result}, Mirage-1 demonstrates performance improvements on the AndroidWorld benchmark, achieving relative gains of $32.3\%$, $23.4\%$, $11.6\%$, and $19.1\%$ in task success rates compared to the OS-Atlas, UGround, Aria-UI, and UI-TARS, respectively. In MobileMiniWob++, Mirage-1 achieves improvements of $19.1\%$, $16.7\%$, and $18.8\%$ over the current SOTA agent with OS-Atlas, UGround, and Aria-UI, respectively. Experimental results in Table \ref{mobile_result} demonstrate that Mirage-1 significantly improves the performance of offline-trained agents in online environments.

\noindent\textbf{Web.}
As shown in Table \ref{mind2web_live_result}, Mirage-1 achieved superior performance across various Operators compared to the baselines. Specifically, we observed improvements in task success rates of $15.8\%$, $15.1\%$, and $8.6\%$ for OS-Atlas, UGround, and Aria-UI, respectively. These consistent improvements across diverse web-based tasks suggest our HMS exhibits strong generalization capabilities in web environments.

The experimental results in Tables \ref{mobile_result} and \ref{mind2web_live_result} demonstrate Mirage-1's cross-platform generalization and its adaptability to various Operators, highlighting the flexibility of the Mirage-1 architecture.

\begin{table}[tbp]
  \caption{Ablation study on different sources of HMS and components of HMS}
  \label{tab:abl}
  \centering
  \begin{subtable}{0.45\textwidth}
    \centering
\resizebox{\textwidth}{!}{
   \begin{tabular}{cc|c}
\toprule[1.1pt]
\multicolumn{2}{c|}{\textbf{Ablation Setting}}     & \multicolumn{1}{c}{\textbf{AndroidWorld}}       \\ \midrule
Offline Skills & Online Skills & SR \\ \midrule
&  & 31.9 \\
&  \checkmark & 34.5 \\
\checkmark & & 37.9  \\
\rowcolor[HTML]{E7EEFE} \checkmark & \checkmark & \textbf{42.2} \\ 
\bottomrule[1.1pt]
\end{tabular}
}
\caption{\label{abl-source}
Ablation study on different sources of HMS}
  \end{subtable}
  \hfill
\begin{subtable}{0.45\textwidth}
    \centering
    \resizebox{0.9\textwidth}{!}{
    \begin{tabular}{ccc|c}
\toprule[1.1pt]
\multicolumn{3}{c|}{\textbf{Ablation Setting}}     & \multicolumn{1}{c}{\textbf{AndroidWorld}}       \\ \midrule
Execution. & Core. & Meta. & SR \\ \midrule
& & & 31.9 \\
\checkmark &  &  & 32.7  \\
\checkmark & \checkmark &  & 35.3  \\
& \checkmark    & \checkmark & 38.8 \\
\rowcolor[HTML]{E7EEFE} \checkmark & \checkmark & \checkmark & \textbf{42.2} \\ \bottomrule[1.1pt] 
\end{tabular}
}
\caption{\label{abl-memory}
Ablation study on the components of HMS}
\end{subtable}
\vspace{-10pt}
\end{table}

\subsection{Long-horizon Benchmark: AndroidLH}
While existing online benchmarks \cite{rawles2024androidworld} focus on short-horizon, single-app tasks, we introduce \textbf{AndroidLH}, an Android benchmark for long-horizon, multi-app operations. AndroidLH includes $30$ diverse tasks across multiple applications, generated via \texttt{GPT-4o} to mirror real-world application scenarios. For evaluation, we adopt AndroidWorld's system state-based completion verification method. See Appendix \ref{appendix-androidlh} for more details.

As shown in Table \ref{mobile_result}, Mirage-1 achieves substantial and consistent improvements across multiple Operators in AndroidLH. Specifically, Mirage-1 exhibits significant performance gains with mean improvements of $53.3\%$ in completion rate and $56.1\%$ in success rate compared to the baseline agents. We attribute this to the fact that HMS provides Mirage-1 with the essential skills required for long-horizon planning in the online environment.

\begin{table}[tbp]
\vspace{-12pt}
  \caption{Experiments about the inference time and efficiency of memory mechanisms}
  \label{tab:other-method}
  \centering
\begin{subtable}[t]{0.45\linewidth}
    \centering
\resizebox{1.2\textwidth}{!}{
\begin{tabular}{lcc}
\toprule[1.1pt]
\textbf{Method} & \textbf{AndroidLH$_{SR}$} & \textbf{Inference time (s) / step} \\ \hline
Mobile-Agent-v2 \citep{wang2024mobile-agent-v2} & 30.0 & 43.5 \\
Mobile-Agent-E \citep{wang2025mobile-e} & 33.3 & 30.1 \\
\rowcolor[HTML]{E7EEFE} Mirage-1 & \textbf{36.7} & \textbf{23.4} \\
\bottomrule[1.1pt]
\end{tabular}
}
\caption{\label {tab-inference-speed} Comparison of inference time}
  \end{subtable}
  \hfill
\begin{subtable}[t]{0.45\linewidth}
    \centering
    \resizebox{0.85\textwidth}{!}{
    \begin{tabular}{lcc|c}
\toprule[1.1pt]
\textbf{Memory Mechanisms} & \textbf{AndroidWorld$_{SR}$} \\ \hline
exemplar memory \citep{zheng2023synapse} &  34.5 \\
workflow \citep{wang2024workflow-memory} & 37.9 \\
\rowcolor[HTML]{E7EEFE} HMS & \textbf{42.2} \\ 
\bottomrule[1.1pt] 
\end{tabular}
}
\caption{\label{tab-memory-method} Comparison of different memory mechanisms}
\end{subtable}
\end{table}

\subsection{Ablation Study}

\label{sec:ablation}
\begin{wrapfigure}[16]{r}{0.49\textwidth}
\centering
\vspace{-12pt}
\includegraphics[width=0.49\textwidth]{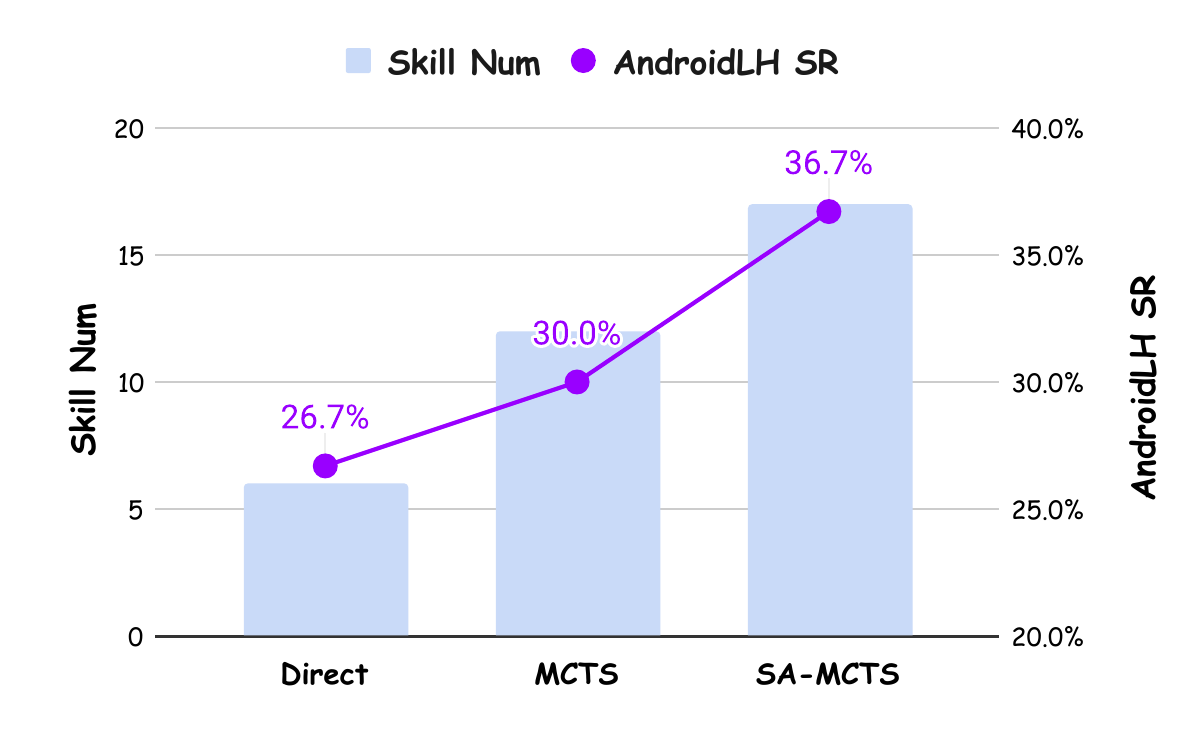}
\vspace{-15pt}
\caption{\small
We compared three online exploration methods: direct exploration, MCTS, and SA-MCTS on $30$ \texttt{GPT-4o} generated tasks. Results show that SA-MCTS acquires more skills than the other methods, demonstrating its superior effectiveness in online exploration.}
\label{abl-mcts}
\end{wrapfigure}
\textbf{Sources of HMS.}  As shown in Table \ref{abl-source}, our ablation experiments demonstrate the relative contributions of different source skills. Removing offline skills led to the most significant performance degradation of $22.3\%$, highlighting their fundamental importance. The subsequent removal of online skills resulted in an $11.3\%$ decrease.  Experimental results reveal that offline skills, serving as the foundation of HMS, provide the necessary knowledge for long-horizon planning. Meanwhile, online skills further improve Mirage-1’s capabilities in online environments.

\noindent\textbf{Components of HMS.}  Table \ref{abl-memory} illustrates the contributions of different types of skills within HMS. When \textit{Execution Skills} were removed, a performance decrease of $8.7\%$ was observed. The removal of \textit{Meta Skills} resulted in a more substantial performance degradation of $19.5\%$, indicating their fundamental role in skill organization and contextual coherence. When \textit{Core Skills} and \textit{Meta Skills} were removed, the performance dropped significantly by 29.1\%. The experimental results indicate that \textit{Core Skills} and \textit{Meta Skills} play a critical role within HMS, providing Mirage-1 with abstract and comprehensive knowledge.

\noindent\textbf{Superiority of HMS}. We compare Mirage-1 with popular agent frameworks in terms of performance and inference efficiency. As shown in the Table \ref{tab-inference-speed}, Mirage-1 outperforms Mobile-Agent-v2 \cite{wang2024mobile-agent-v2} and Mobile-Agent-E \cite{wang2025mobile-e} on AndroidLH, while also achieving faster inference speed. We attribute this to Mirage-1 benefiting from the HMS module, which enables efficient retrieval and planning, rather than performing multi-round planning and decision-making at every step like other agents. Moreover, we compare the proposed HMS module with other memory mechanisms. As shown in Table \ref{tab-memory-method}, experimental results demonstrate that HMS outperforms existing memory mechanisms in the AndroidWorld benchmark. 

\noindent\textbf{Online Exploration Strategy.} As shown in Figure \ref{abl-mcts}, the SA-MCTS achieved remarkable improvements in skill acquisition efficiency, demonstrating a $2.8\times$ increase compared to the Direct approach and a $41\%$ improvement over standard MCTS. The results show that HMS can significantly improve the success rate of MCTS exploration, demonstrating the strong adaptability of HMS.

\begin{figure*}[tbp]
    \centering
    \includegraphics[width=1.0\textwidth]{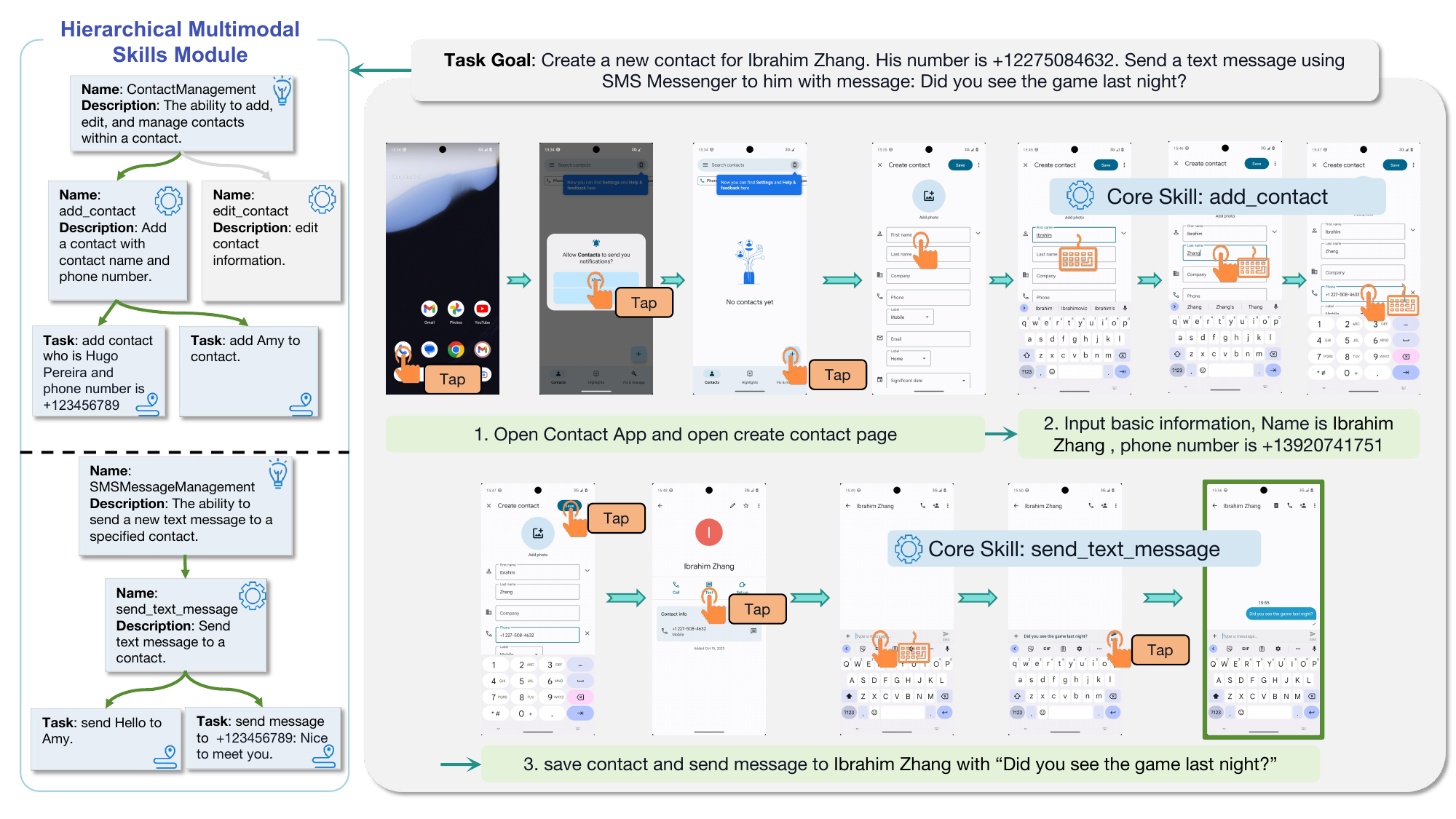}
    \caption{Case study example where Mirage-1 completes a long-horizon task. First, Mirage-1 retrieval Meta Skills, Core Skills, and Execution Skills from HMS. Then Mirage-1 generates plans for long-horizon tasks. With the help of Core Skills, Mirage-1 can easily achieve sub-goals.}
    \label{fig-case_study}
\end{figure*}
\subsection{Case Study}

To show the Mirage-1's performance in executing long-horizon tasks in real-world scenarios, we conducted a case study, as shown in Figure \ref{fig-case_study}.
To accomplish the task goal, Mirage-1 first needs to create a new contact and accurately fill in the contact’s information. It then must also learn how to send a message to the specified contact. The qualitative case in Figure \ref{fig-case_study} shows that Mirage-1 effectively retrieves and leverages skills in HMS to complete the long-horizon task.

\section{Conclusion}
In this paper, we propose a Hierarchical Multimodal Skills module (HMS) that addresses the challenge of insufficient prior knowledge in long-horizon task planning. To address the domain gap between offline and online, a Skill-Augmented Monte Carlo Tree Search (SA-MCTS) algorithm is proposed. This algorithm effectively utilizes offline-acquired skills to reduce the action search space during online tree exploration. On top of HMS, we propose multimodal agent Mirage-1. Experimental results demonstrate that Mirage-1 achieves superior performance compared to SOTA GUI agents, particularly in long-horizon tasks.

\bibliography{custom}
\bibliographystyle{plainnat}
\clearpage
\appendix

\section{Limitations}
Multimodal large language models (MLLMs) possess inherent strengths and limitations \cite{chen2024lion,shen2024mome}. Leveraging their capabilities in instruction understanding and multimodal perception \cite{li2025optimus}, agents have achieved promising results across various domains \cite{li2025lion,li2025star}. However, issues related to the interpretability of their outputs and potential safety risks remain significant concerns that cannot be overlooked. Moreover, although demonstrating promising results on the online evaluation benchmark, the current implementation exhibits computational overhead in inference processing. A critical architectural limitation exists in the integration of planner and grounding components: the grounding model's interpretation accuracy of planner outputs directly impacts task execution success. Misinterpretation of planning results by the grounding model can lead to imprecise spatial positioning and subsequent task failures. 

\section{Broader Impact}

The large language model exhibits an inherently uncontrollable output in the planning, action generation, and reflection phases, potentially producing outputs that violate safety constraints or ethical guidelines. Our evaluation framework is implemented within web and mobile environments. Furthermore, model inputs may contain sensitive user information, introducing potential privacy vulnerabilities in inference time.

\section{GUI environment}
\label{sec:appendix-env}
\subsection{Mobile}
Following the settings of AndroidWorld \cite{rawles2024androidworld}, we create an Android emulator with the hardware set as Pixel 6, the System Image as Tiramisu, and the API level as 33. AndroidWorld determines task completion by utilizing predefined evaluation functions that assess system states, including database records and device information. This automated verification mechanism uses system-level indicators to validate the execution status of the task.

\noindent\textbf{Observation Space.}
AndroidWorld includes three observation types: Screenshots, Accessibility tree, and UI elements. Mirage-1 only receives the current screenshot of the device. The screenshot resolution is $2400 \times 1080 \times 3$.

\noindent\textbf{Action Space.} The action space of Mirage-1 includes:
\begin{itemize}
    \item Click: Simulates touch events as specified coordinates
    \item Text input: Simulates typing in focused text fields
    \item Navigation: Sends navigate home/ navigate back key events
    \item Scrolling: Executes swipes in four directions(up, down, left, right)
    \item App launching: Starts specified applications
    \item Status: Reports if the task is in progress, complete, or infeasible.
    \item Answer: Provides responses, which are needed for information retrieval tasks
    \item Wait: No-op useful for loading screens and UI transitions
    \item Clear Text: helpful operation to clear text in text fields
\end{itemize}

\subsection{Web}
Following the settings of Mind2Web-Live \cite{pan2024webcanvas}, we use playwright \footnote{\url{https://github.com/microsoft/playwright}} to simulate the web environment. Mind2Web-Live validates task completion through predetermined key nodes that are guaranteed to execute. The system evaluates the current task status by analyzing both the active URL address and the interaction metadata of clicked elements.

\noindent\textbf{Observation Space.} Mind2Web-Live includes two observation types: Screenshots and Accessibility tree. Mirage-1 only receives current webpage capture. The screenshot resolution is $1080 \times 720 \times 3$.

\noindent\textbf{Action Space.} The action space of Mirage-1 includes:
\begin{itemize}
    \item Goto: Goto specific URL
    \item Google Search: use Google to search content with keywords
    \item Click: Mouse click event with specified coordinates
    \item Hover: Mouse hover event with specified coordinates
    \item Fill: Typing text in focused text fields
    \item Go Back: Go to the previous webpage
\end{itemize}

\section{AndroidLH}
\label{appendix-androidlh}
AndroidLH comprises 30 long-horizon tasks spanning 12 applications. As shown in Table \ref{long_horizon_benchmark}, the table below lists all tasks. We have leveraged AndroidWorld's task template and evaluation mechanisms to create logical combinations of different tasks that reflect real-world usage scenarios. 

For evaluation, we adopt AndroidWorld's system state-based completion verification method while introducing an additional metric, \textit{Completion Rate} (CR), which measures the proportion of completed subtasks. This metric provides finer-grained insights into task progress, particularly for complex, multi-step tasks.

\section{SA-MCTS}
\label{appendix-sa-mcts}
SA-MCTS consists of four phases: selection, expansion, simulation, and backpropagation.

\subsection{Selection}
The selection phase uses the Upper Confidence Bound (UCB1) formulation of MCTS to select nodes which aims to balance exploration and exploitation. We denote the agent state with $h_t$. We consider the value function $Q(h_t, g)$ which represents the estimated value of executing sub-goal $g$ at state $h_t$. At each new node $l_t$, we sample $S$ proposal sub-goals $g_t^1, \dots, g_t^S$. We initialize all values $Q(h_t, g)$ to zero. We use use \texttt{GPT-4o} to produce a feedback score for each sub-goal by asking it to rank the generated sub-goals by its perceived utility in
helping the agent complete the user task.
After the initial selection, we select the sub-goal to explore based on the standard MCTS UCB1 formulation:
\begin{equation}
    l^*_t=\arg\max_{l_t^1,\dots,l_t^S}\Bigg [Q(h_t,g)+c_{\exp}\sqrt{\frac{\log N(h_t)}{1+N(h_{t+1)}}} \Bigg ]
\end{equation}
where $N(h_t)$ is the visitation frequency of state $h_t$, and $c_{\exp}$ is an exploration constant.

\subsection{Expansion and Simulation}
Based on the preceding section, we select and execute sub-goals in the environment to reach
a new node. Beginning from the selected state node’s trace, we roll out the trajectory until a terminal state is reached. The environment returns a reward at the end of the trajectory, $R$, where $R=1$ if the agent completes the sub-goal and $R=0$ otherwise.

\subsection{Backpropagation}
We then backpropagate this reward by updating the values of each node bottom-up from the leaf node to the root as follows:
\begin{equation}
Q(h_t,g_t^i) \leftarrow \frac{Q(h_t,g_t^i)N(h_t,g_t^i)+R}{N(h_t,g_t^i)+1}
\end{equation}
\begin{equation}
    N(h_t,g_t^i) \leftarrow N(h_t,g_t^i) +1
\end{equation}

\section{Experiments}
\subsection{Experimental Settings}
\label{appendix-exp-setting}
To demonstrate the generalization of Mirage-1 across diverse domains, we evaluated it both in mobile and web online environments following previous work (\cite{wang2024mobile-agent}, \cite{wang2024mobile-agent-v2}, \cite{yang2023appagent}, \cite{he-etal-2024-webvoyager}). To further verify the effectiveness of the method, we selected various grounding models, including OS-Atlas \cite{wu2024osatlas}, UGround \cite{gou2024uground}, Aria-UI \cite{yang2024aria-ui}, and UI-TARS \cite{qin2025ui-tars}. All experiments were conducted on 8x NVIDIA L40 GPUs.

\noindent\textbf{Mobile.} We evaluate on AndroidWorld \cite{rawles2024androidworld}, an online benchmark within an Android emulator environment. It consists of 116 tasks that span 20 applications and assesses the success or failure of the task by examining the final state of the virtual device's system. 
We also include 92 MobileMiniWob++ tasks provided by AndroidWorld, which adapts the MiniWob++ Web agent environment \cite{liu2018miniwobplusplus} to AndroidEnv \cite{toyama2021androidenv}, the same environment as AndroidWorld. 
We report \textit{task success rate} as an evaluation metric, which represents the proportion of successfully completed tasks relative to the total number of tasks.

\noindent\textbf{Web.} We utilize the test set from Mind2Web-Live \cite{pan2024webcanvas}, an online evaluation benchmark designed for dynamic web environments. It includes 104 tasks and adds evaluation functions that automatically assess task success or failure. Specifically, it defines and annotates key nodes for each task, which represent critical steps that must be completed for a task to be deemed successful, regardless of the trajectory taken by the agent. We use standard metrics: \textit{micro completion rate}, which measures the proportion of completed key nodes across all the tasks, and \textit{task success rate}, which represents the proportion of successfully completed tasks relative to the total number of tasks.

\subsection{Baselines}
\label{appendix-exp-baseline}
\textbf{Mobile.} We conducted a comprehensive comparative analysis of multiple agents and their variants across different input modalities. 
For AXTree-only input, we compare the performance of GPT-4-Turbo, GPT-4o, and Gemini 1.5 Pro as operators that use choice-based element grounding.
In the combined Image and AXTree scenario, we examine GPT-4-Turbo, Gemini 1.5 Pro, GPT-4o, and SeeAct, all implementing SoM grounding methods \cite{yang2023setofmark}.
For image-only inputs, we mainly removed SoM images and a textual list of elements from the AXTree in the observation and compared GPT-4o with different grounding models (OS-Atlas, UGround, Aria-UI, UI-TARS). All agents employed a ReAct-style reasoning process \cite{yao2022react} for target element selection or localization, enhanced by self-reflection \cite{shinn2024reflexion} methodology after each execution step to optimize subsequent decision-making.

\noindent\textbf{Web.} We evaluate agents in different configurations for web interaction tasks. For text-based inputs, we test the original WebCanvas agent \cite{pan2024webcanvas} with multiple language models including GPT-4-Turbo, Gemini 1.5 Pro, and GPT-3.5-Turbo, where the agent perceives web pages through textual HTML elements with choice-based selection. For image-only inputs, we remove DOM tree components from the original prompts to work purely with visual inputs. We evaluated the performance of GPT-4o with different grounding models (OS-Atlas, UGround, Aria-UI). In all image-based scenarios, the agents interact with HTML elements using coordinate-based localization rather than DOM references.

\subsection{Implementation Details}
\label{appendix-exp-impl}
\textbf{Mobile.} We build Mirage-1 based on M3A Agent \cite{rawles2024androidworld}. We mainly removed SoM images and a textual list of elements from the AXTree in the observation. Only tasks and screenshots are accepted as input. 
We used GPT-4o as the planning and reflection model. Separately, we used OS-Atlas, UGround, Aria-UI, and UI-TARS as our grounding models. 

To initiate HMS, we randomly select the 1000 trajectories in the general data of AITW \cite{rawles2024aitw}. We use the method discussed in Section \ref{sec:hms}.
To expand HMS, we provide GPT-4o with information about the app, including the developer's description of the app's functions and several app screenshots. Then we prompt \texttt{GPT-4o} to generate 200 unseen tasks. After that, we use the SA-MCTS algorithm for online exploration and utilize the successful trajectories to expand our HMS. Finally, we obtained 93 Meta Skills, 463 Core Skills, and 1063 Execution Skills.

\noindent\textbf{Web.} We build Mirage-1 based on the WebCanvas agent \cite{pan2024webcanvas}. We mainly removed the textual list of elements from the DOM tree in the observation and only used webpage screenshots as inputs. Like Mobile, we used GPT-4o as the planning and reflection models and used OS-Atlas, UGround, and Aria-UI as our grounding models. 

To initiate HMS, we randomly select the 1000 trajectories of Multimodal-Mind2Web \cite{zheng2024seeact}, which is the multimodal version of Mind2Web \cite{deng2024mind2web}. We use the method discussed in Section \ref{sec:hms}.
To expand HMS, similar to Mobile, we provide GPT-4o with several web page screenshots and the main content of the web pages. Then, we prompt \texttt{GPT-4o} to generate 200 unseen task descriptions for the web pages. We then SA-MCTS algorithm for exploration and then utilize the successful trajectories to expand our HMS. Finally, we obtained 125 Meta Skills, 643 Core Skills, and 1049 Execution Skills.

\section{Case Study}
\begin{figure*}
    \includegraphics[width=1.0\textwidth]{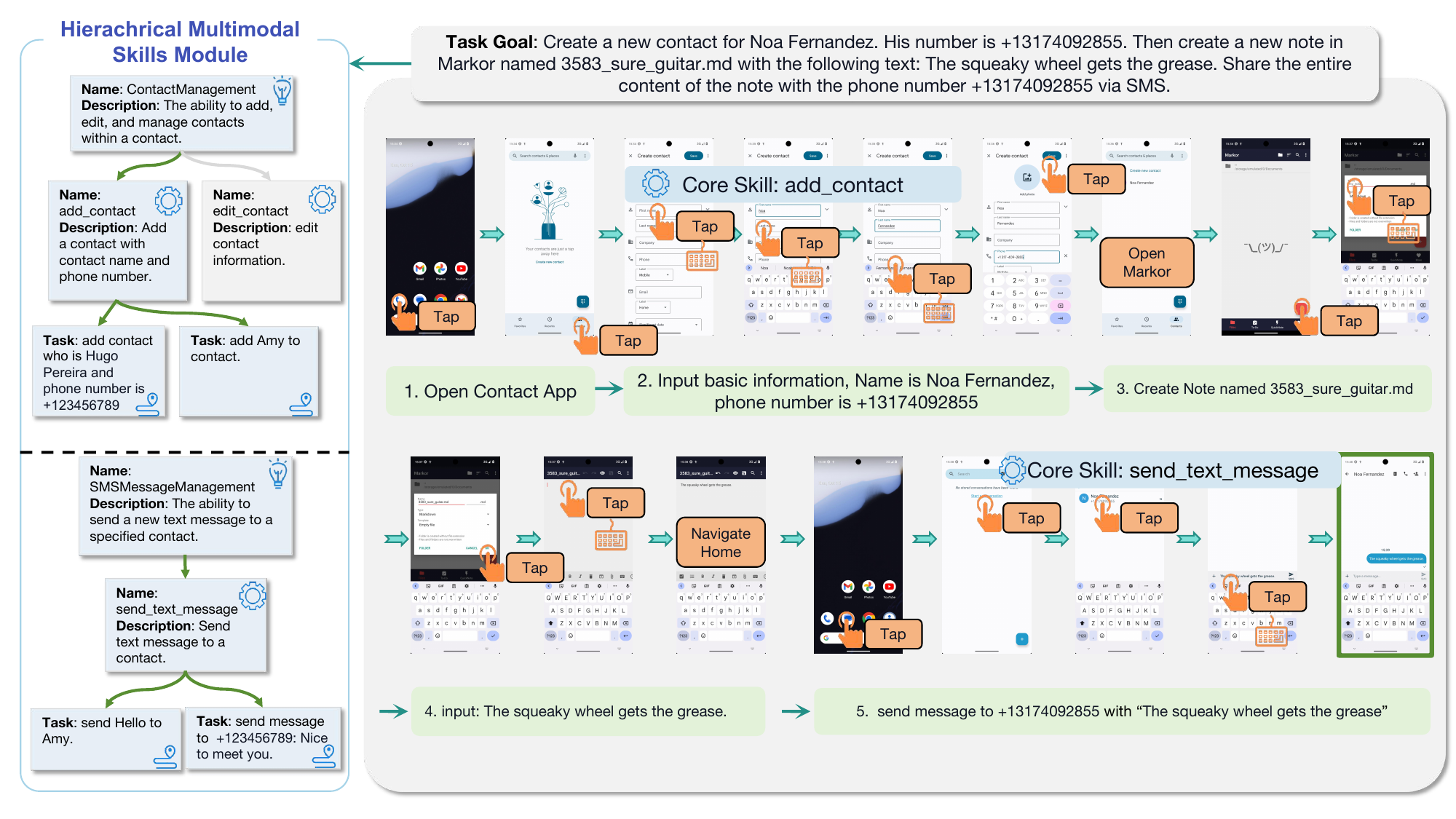}
    \caption{Case study example where Mirage completes a long-horizon task. First, Mirage retrieval Meta Skills, Core Skills, and Execution Skills from HMS. Then Mirage generates plans for long-horizon tasks. With the help of Core Skill, Mirage can easily achieve sub-goals.}
    \label{fig-case_study_appendix}
\end{figure*}

We present a more complex example. As shown in \ref{fig-case_study_appendix}, after creating a contact, the agent needs to create a note. After entering the text, the text should then be sent to someone via text message. Mirage-1 first retrieves HMS and gets 2 relevant core skills: \texttt{add\_contact} and \texttt{send\_text\_message}. Then Mirage-1 prompts \texttt{GPT-4o} to generate plans for the task.

\section{Prompts}

\definecolor{codeblue}{rgb}{0.25,0.5,0.5}
\definecolor{codekw}{rgb}{0.85, 0.18, 0.50}
\definecolor{keywordgreen}{rgb}{0,0.6,0}
\lstset{
  backgroundcolor=\color{gray!10},
  basicstyle=\fontsize{8pt}{9pt}\ttfamily\selectfont,
  columns=fullflexible,
  breaklines=true,
  captionpos=b,
  commentstyle=\fontsize{6.5pt}{7.5pt}\color{codeblue},
  caption={Prompt for Extract Execution Skill from Trajectory.},
  label={lst:trace_skill}
}

\begin{lstlisting} % [language=python]  

You are a skilled assistant specializing in analyzing and interpreting tasks on mobile devices.

Based on user's goal, agent's action and two screenshots (before and after the action), you need to transform the action excuted by agent in the following action list. You need to output the action in the correct JSON format. Note you must describe the goal of action in the 'description' field of the JSON format.

You can perform the following actions:
{action space}

I will give some examples to help you understand the task better:
Example 1:
{example1}

Example 2:
{example2}

Now it's your turn.
Input:
- Goal: {task}.
- Action: {action}.
- Screenshots: Two images before and after the action.

your answer should look like:
Action: {{"action_type": ..., "description": ...}}

Your Answer:

\end{lstlisting}
\definecolor{codeblue}{rgb}{0.25,0.5,0.5}
\definecolor{codekw}{rgb}{0.85, 0.18, 0.50}
\definecolor{keywordgreen}{rgb}{0,0.6,0}
\lstset{
  backgroundcolor=\color{gray!10},
  basicstyle=\fontsize{8pt}{9pt}\ttfamily\selectfont,
  columns=fullflexible,
  breaklines=true,
  captionpos=b,
  commentstyle=\fontsize{6.5pt}{7.5pt}\color{codeblue},
  caption={Prompt for Generate New Core Skill.},
  label={lst:core_skill}
}

\begin{lstlisting} % [language=python]  

Create a new skill function using a given task, action sequence, and existing skill functions. Ensure arguments and actions are relevant and general enough for reuse.
The skill function only supportes these actions:
- `Agent.click(element)`: Click on an element on the screen.
- `Agent.double_tap(element)`: Double tap on an element on the screen.
- `Agent.long_press(element)`: Long press on an element on the screen.
- `Agent.input_text(text)`: Type text into a text field.
- `Agent.scroll(direction)`: Scroll the screen or a scrollable UI element in one of the four directions.
- `Agent.swipe(direction)`: Swipe the screen in one of the four directions.
- `Agent.open_app(app_name)`: Open an app.
- `Agent.wait()`: Wait for the screen to update.
- `Agent.keyboard_enter()`: Press the Enter key.
- `Agent.navigate_home()`: Navigate to the home screen.
- `Agent.navigate_back()`: Navigate back.
Tips:
- Use `Agent.open_app(app_name)` if app opening is a primary initial step.
# Steps
1. Analyze the provided task and the sequence of actions to understand the goal.
2. Reference existing skill functions as templates for creating new ones.
3. Generate a skill function, ensuring its actions and arguments match the provided task and actions.
4. Ensure the skill function's arguments are reusable for similar tasks.
# Output Format
- Provide the new skill function in Python format.
- Ensure the skill function includes a docstring describing its parameters and purpose.
# Examples
{example1}
{example2}
{example3}
Your Turn:
Input:
- Task: {task}
- Actions:
{actions}
- Skill: {skill}
Your Answer:
Reason: <reason>
New skill: <new skill>

\end{lstlisting}
\definecolor{codeblue}{rgb}{0.25,0.5,0.5}
\definecolor{codekw}{rgb}{0.85, 0.18, 0.50}
\definecolor{keywordgreen}{rgb}{0,0.6,0}
\lstset{
  backgroundcolor=\color{gray!10},
  basicstyle=\fontsize{8pt}{9pt}\ttfamily\selectfont,
  columns=fullflexible,
  breaklines=true,
  captionpos=b,
  commentstyle=\fontsize{6.5pt}{7.5pt}\color{codeblue},
  caption={Prompt for Generate Select or Generate Meta Skill.},
  label={lst:meta_skill}
}

\begin{lstlisting} % [language=python]  

You are an expert in abstract thinking and plan optimization. Your task is to classify a given task into a suitable category based on a list of existing skills. Each skill is defined by its name and description in the format `skill_name: skill_description`.
- Identify and Match: Start by thoroughly analyzing the task and the available skills. Match the task with the skill descriptions to determine the most appropriate category.
- Create New Skills: If no existing skill entirely covers the task, synthesize a new skill by combining existing skills or creating a completely new description that captures the necessary abilities.
- Consider Multiple Skills: Be aware that some tasks may require a combination of skills, so consider all relevant skills when classifying the task.

### New Skill Template
Skill name: Provide a concise, descriptive name for the skill.
Skill description: Write a detailed description that covers the essence of the skill.
Skill combination: If applicable, list the combined skills in the format `<skill1>, <skill2>, <skill3>`.

### Steps
1. Analyze the Task: Start by considering the main actions or objectives involved in the given task.
2. Review Existing Skills: Go through each skill description to find potential matches or overlaps with the task requirements.
3. Decide on a Category: Determine whether the task can be fully categorized under an existing skill, needs the combination of several skills, or demands the creation of a new skill.
4. Define and Document: Clearly write out the reasoning and final categorization, including new skill descriptions if necessary.

### Output Format
Reason: Provide the thought process and analysis that led to the classification.
Category: Specify the appropriate existing skill or indicate "New Skill" with details.

### Examples
{example1}
{example2}
{example3}
{example4}
{example5}

### Notes
- Consider potential overlaps between skills.
- Always detail your decision-making process clearly.
- Generate new skills only when necessary.

Input:
Task: {task}
Skills:
{skills}

Output:
Reason: <reason>
Category: <category>
Skill name: <skill name>(if new skill)
Skill description: <skill description>(if new skill)
Skill combination: <skill combination>(if new skill)

\end{lstlisting}
\definecolor{codeblue}{rgb}{0.25,0.5,0.5}
\definecolor{codekw}{rgb}{0.85, 0.18, 0.50}
\definecolor{keywordgreen}{rgb}{0,0.6,0}
\lstset{
  backgroundcolor=\color{gray!10},
  basicstyle=\fontsize{8pt}{9pt}\ttfamily\selectfont,
  columns=fullflexible,
  breaklines=true,
  captionpos=b,
  commentstyle=\fontsize{6.5pt}{7.5pt}\color{codeblue},
  caption={Prompt for Planning with HMS.},
  label={lst:plan}
}

\begin{lstlisting} % [language=python]  

You are an Android Phone Agent that generates high-level task plans. Given a task and skill functions, generate plans in either of these formats:
- Call skill function with specific arguments: `Category.skill_function(arg1, arg2, ...)`.
- Use natural language to describe the plan.  You should use this format: In order to complete <goal>, I need to <do something> in <some context>. Replace <goal> with the specific goal, <do something> with the high-level task, and <some context> with the relevant information or condition.

There are some nesseray rules for you to follow:
- When using natural language, the description should be clear and easy to understand.
- Write code in a single line without line breaks
- Only use provided skill functions, do not create new ones
- Use natural language for tasks that don't match existing functions
- Ensure arguments match function specifications exactly
- DO not add reasoning process in the Plans.

Example:
Task: "Create a note in Markor named abcs.txt. Perform a paste operation in the note and save the note."
Category: NoteManager
Skill Functions:
def add_text_to_bottom(note_name, additional_text):
    \"\"\"Edit a specified note in the Markor app by adding text to the bottom.

    Args:
        note_name (str): The name of the note to be edited.
        additional_text (str): The text to add to the bottom of the note.
    \"\"\"

def create_note_in_markor(note_name, note_content):
    \"\"\"Create a new note in the Markor app with the specified name and content.

    Args:
        note_name (str): The name of the note to be created.
        note_content (str): The content to be added to the new note.
    \"\"\"

Result:
Reason: The user wants to create a note in Markor named abcs.txt, perform a paste operation in the note, and save the note. I don't need to call the skill function `add_text_to_bottom` because I has performed the paste operation in the note.
Plans:
1. NoteManager.create_note_in_markor("abcs.txt", "")
2. Perform paste operation in the note.
3. Check the note and save the note.

Think carefully about the task and skill functions to generate the most accurate and feasible plans.

Input:
Task: {task}
{context}
Output must be in the following format:
Reason: <reason>
Plans:
1. plan1
2. plan2

\end{lstlisting}
\definecolor{codeblue}{rgb}{0.25,0.5,0.5}
\definecolor{codekw}{rgb}{0.85, 0.18, 0.50}
\definecolor{keywordgreen}{rgb}{0,0.6,0}
\lstset{
  backgroundcolor=\color{gray!10},
  basicstyle=\fontsize{8pt}{9pt}\ttfamily\selectfont,
  columns=fullflexible,
  breaklines=true,
  captionpos=b,
  commentstyle=\fontsize{6.5pt}{7.5pt}\color{codeblue},
  caption={Prompt for Reflection with Execution Skill.},
  label={lst:reflection}
}

\begin{lstlisting} % [language=python]  

You are an instructor in the Android Phone scenario, guiding the Agent to better complete users' tasks. Given the user's task, the current screenshot of the Android Phone, and an action that the Agent intends to take, assess whether the action can help to complete the goal. Predict the new state that the action may lead to, based on your previous experience, and assign a score to each action. The Agent will then choose the action with the highest score to execute.

I will provide examples to help you predict the new state.
{exampl_trace_skill1}
{exampl_trace_skill2}
{exampl_trace_skill3}

Input:
- The current user goal is: {goal}
- The current action that the Agent intends to take is: {action}
- The summary of the Agent's historical trajectory is as follows: {history}
- I will give the first screenshot and current screenshot of the Android Phone to help you predict the new state that action may lead to based on your previous experience.

Think step by step to predict the new state, evaluate if the action aligns with the goal, and rank the action accordingly.

Output:
You must output the structured answer based on the above information.
- caption: Describe the screenshot using less than 30 words.
- reason: Think step by step to predict the new state and evaluate if the proposed action aligns with the goal based on your previous experience.
- state_change: Predict the new state that action may lead to based on your previous experience. Output only the state_change.
- score: Assign a score (0-10) to the action based on the following criteria:
  1) Goal Alignment: Does the action help achieve the user's goal?
  2) How well the action brings the user closer to their goal.
  3) Likelihood of Success: How likely is it that the action will succeed based on the current context and UI state.
  4) Efficiency: How efficient is the action in terms of user effort and time required.
  just output the score as a number between 0-10.

Output format:
{{"caption": ..., "reason": ..., "state_change": ..., "score": ...}}

\end{lstlisting}

\begin{table*}
\centering
\scalebox{0.6}{
\begin{tabular}{p{0.3\textwidth}<{\centering}p{0.9\textwidth}<{\centering}p{0.2\textwidth}<{\centering}}
\toprule[1.5pt]

\textbf{Name} & \textbf{Task Template} &  \textbf{Apps} \\
\midrule
AddRecipeAndCleanDuplicates & Add the following recipes into the Broccoli app:\{recipe\}. Then delete all but one of any recipes in the Broccoli app that are exact duplicates, ensuring at least one instance of each unique recipe remains.  & broccoli app \\ \midrule
OptimizeBrightnessForPhoto & Turn the screen brightness up to the maximum to ensure the shooting environment is well-lit, then take a photo.  & settings, camera \\ \midrule
OptimizeBrightnessForVideo & Turn the screen brightness up to the maximum to ensure the shooting environment is well-lit, then take a video.  & settings, camera \\ \midrule
SimpleCalendarAddTomorrowAnd DeleteEvents & In Simple Calendar Pro, create a calendar event for tomorrow at \{hour\}h with the title '\{event\_title\}' and the description '\{event\_description\}'. The event should last for \{duration\_mins\} mins. Then, delete all events in the calendar on \{year\}-\{month\}-\{day\}.  & simple calendar pro \\ \midrule
SimpleCalendarDeleteEvents OnRelativeDayAndAddOneEventRelativeDay & In Simple Calendar Pro, delete all events scheduled for this \{day\_of\_week\}.Then create a calendar event for this \{day\_of\_week\} at \{hour\}h with the title '\{event\_title\}' and the description '\{event\_description\}'. The event should last for \{duration\_mins\} mins.  & simple calendar pro \\ \midrule
ClockStopWatchRunning
AndBrowserDraw & Run the stopwatch. Open the file task.html in Downloads in the file manager; when prompted open it with Chrome. Then create a drawing using the three colors shown at the top and hit submit.  & chrome, clock \\ \midrule
ClockStopWatchRunning
AndBrowserMaze & Run the stopwatch. Open the file task.html in Downloads in the file manager; when prompted open it with Chrome. Then navigate the X to the bottom-right cell, by using the" " direction buttons.  & chrome, clock \\ \midrule
ClockStopWatchRunning
AndBrowserMultiply & Run the stopwatch.  Then click the button 5 times, remember the numbers displayed, and enter their product in the form.  & chrome, clock \\ \midrule
ClockStopWatchRunning AndBrowserMazeAndPause & Run the stopwatch. Open the file task.html in Downloads in the file manager; when prompted open it with Chrome. Then navigate the X to the bottom-right cell, by using the" " direction buttons. After that, pause the stopwatch.  & chrome, clock \\ \midrule
ClockStopWatchRunning AndBrowserDrawAndPause & Run the stopwatch. Open the file task.html in Downloads in the file manager; when prompted open it with Chrome. Then create a drawing using the three colors shown at the top and hit submit. After that, pause the stopwatch.  & chrome, clock \\ \midrule
ClockStopWatchRunningAnd
BrowserMultiplyAndPause & Run the stopwatch.  Then click the button 5 times, remember the numbers displayed, and enter their product in the form. After that, pause the stopwatch.  & chrome, clock \\ \midrule
ClockStopWatchRunningAndPause & Run the stopwatch. And pause it after a while.  & clock \\ \midrule
ConnectBluetoothAndOpenMusic & Turn on bluetooth. Then open the retro music app. Clear any pop-ups that may appear by granting" " all permissions that are required. Enjoy the music!  & settings, retro music \\ \midrule
AddContactAndCallAndSms & Create a new contact for \{name\}. His number is \{number\}. Then call him. After that, send a text message to him using Simple SMS Messenger: nice to meet to you.  & contacts, dialer, simple sms messenger \\ \midrule
AddContactAndCall & Create a new contact for \{name\}. His number is \{number\}. Then call him.  & contacts, dialer \\ \midrule
AddContactAndMarkowAndSms & Create a new contact for \{name\}. His number is \{number\}. Then create a new note in Markor named \{file\_name\} with the following text: \{text\}. Share the entire content of the note with the phone number \{number\} via SMS using Simple SMS Messenger.  & contacts, markor, simple sms messenger \\ \midrule
AddContactAndSms & Create a new contact for \{name\}. His number is \{number\}. Send a text message using Simple SMS Messenger to him with message: \{text\}.  & contacts, simple sms messenger \\ \midrule
CopyToClipboardAndCreateNote & Copy the following text to the clipboard: \{clipboard\_content\}. Then create a note in Markor named \{file\_name\}. Perform a paste operation in the note and save the note.  & markor \\ \midrule
SystemCopyToClipboardAndSms & Copy the text \{clipboard\_content\} to your clipboard, then send it to \{number\} using simple sms messenger app.  & simple sms messenger \\ \midrule
MarkorCreateFolderAndMoveFile & Create a new folder in Markor named \{folder\_name\}.Then move the note \{file\_name\} from \{source\_folder\} to \{destination\_folder\}.  & markor \\ \midrule
MarkorCreateNoteAndSms & Create a new note in Markor named \{file\_name\} with the following text: \{text\}. Share the entire content of the note with the phone number \{number\} via SMS using Simple SMS Messenger  & markor, simple sms messenger \\ \midrule
TurnOffWifiAndAudioRecorder & Turning off Wifi. Then Record an audio clip using Audio Recorder app and save it.  & settings, audio recorder \\ \midrule
TurnOffWifiAnd
ConnectBluetooth
AndOpenMusic & Turn off wifi and turn on bluetooth. Then open the retro music app. Clear any pop-ups that may appear by granting all permissions that are required. Enjoy the music!  & settings, retro music \\ \midrule
TurnOffWifiAnd
TurnOffBluetooth
AudioRecorder & Turning off Wifi and turn off bluetooth. Then Record an audio clip using Audio Recorder app and save it.  & settings, audio recorder \\ \midrule
TurnOnWifiAndBrowserMaze & Turning on Wi-Fi. Open the file task.html in Downloads in the file manager; when prompted open it with Chrome. Then navigate the X to the bottom-right cell, by using the direction buttons.  & settings, chrome \\ \midrule
CameraTaskPhotoAndCreate
Drawing & Take a photo with the camera. Then, Create a new drawing in Simple Draw Pro. Name it \{file\_name\}. Save it in the Pictures folder within the sdk\_gphone\_x86\_64 storage area.  & camera, simple draw pro \\ \midrule
TurnOnWifiAndOpenApp & Turn on Wifi, then open the \{app\_name\} app  & settings \\ \midrule
TurnOffWifiAndTurnOnBluetooth & Turn off WiFi, then enable bluetooth  & settings \\ \midrule
TurnOffWifiAndTurnOffBluetooth & Turn off WiFi, then turn off bluetooth  & settings \\ \midrule
UpdateExpensesWithCleanup & Add the following expenses into the arduia pro expense: \{expense\}. Then, delete all but one of any expenses in arduia pro expense that are exact duplicates, ensuring at least one instance of each unique expense remains.  & pro expense \\
\bottomrule[1.5pt]
\end{tabular}
}
\caption{\label{long_horizon_benchmark}
AndroidLH Benchmark Task list.
}
\end{table*}

\end{document}